\documentclass[letterpaper, 10 pt, conference]{ieeeconf}  

\let\labelindent\relax
\usepackage{enumitem}

\IEEEoverridecommandlockouts              
\usepackage{times}
\usepackage{epsfig}
\usepackage{graphicx}
\usepackage{amsmath}
\usepackage{amssymb}
\usepackage{authblk}
\usepackage{graphicx} 
\usepackage{dblfloatfix}  
\usepackage{multirow}
\usepackage{booktabs} 
\usepackage{enumitem} 
\usepackage{subcaption} 
\usepackage{arydshln}
\usepackage{float}

\usepackage[pagebackref=true,breaklinks=true,letterpaper=true,colorlinks,bookmarks=false,linkcolor=black, citecolor=black, urlcolor=black]{hyperref}

\title{\LARGE \bf V2X-DGPE: Addressing Domain Gaps and Pose Errors for Robust Collaborative 3D Object Detection*}

\author{
    Sichao Wang$^{1}$, Ming Yuan$^{1}$, Chuang Zhang$^{1}$, Lei He$^{1}$, Qing Xu$^{1}$, Jianqiang Wang$^{1}$\\
    \thanks{*This work was supported by the National Natural Science Foundation of China, Science Fund for Creative Research Groups (Grant No. 52221005).}
    \thanks{$^{1}$Sichao Wang, Ming Yuan, Chuang Zhang, Lei He, Qing Xu, and Jianqiang Wang are with the School of Vehicle and Mobility, Tsinghua University, Beijing, 100084, China. {\tt\small wjqlws@tsinghua.edu.cn}}%
}

\begin{document}

\maketitle
\thispagestyle{empty}
\pagestyle{empty}

\begin{abstract}   
In V2X collaborative perception, the domain gaps between heterogeneous nodes pose a significant challenge for effective information fusion. Pose errors arising from latency and GPS localization noise further exacerbate the issue by leading to feature misalignment. To overcome these challenges, we propose V2X-DGPE, a high-accuracy and robust V2X feature-level collaborative perception framework. V2X-DGPE employs a Knowledge Distillation Framework and a Feature Compensation Module to learn domain-invariant representations from multi-source data, effectively reducing the feature distribution gap between vehicles and roadside infrastructure. Historical information is utilized to provide the model with a more comprehensive understanding of the current scene. Furthermore, a Collaborative Fusion Module leverages a heterogeneous self-attention mechanism to extract and integrate heterogeneous representations from vehicles and infrastructure. To address pose errors, V2X-DGPE introduces a deformable attention mechanism, enabling the model to adaptively focus on critical parts of the input features by dynamically offsetting sampling points. Extensive experiments on the real-world DAIR-V2X dataset demonstrate that the proposed method outperforms existing approaches, achieving state-of-the-art detection performance. The code is available at https://github.com/wangsch10/V2X-DGPE.
\end{abstract}

\section{INTRODUCTION}
Recent works have contributed many high-quality collaborative perception datasets such as V2X-Sim \cite{li2022v2xsim}, V2X-Set \cite{v2xvitxu2022v2x}, Dair-V2X  \cite{yu2022dair}, OPV2V  \cite{xu2022opv2v}, etc. Current collaborative perception methods \cite{chen2019cooper,wang2020v2vnet,xu2022opv2v} largely rely on vehicle-to-vehicle (V2V) communication, overlooking roadside infrastructure. Asynchronous triggering and transmission between vehicle and infrastructure sensors introduce delays \cite{lei2022latency}, while GPS positioning noise exacerbates sensor lag and coordinate transformation errors. This results in serious spatial-temporal errors. While existing algorithms have made progress in addressing these issues, they often fail to meet practical application demands. Therefore, there is an urgent need to solve the problem of feature misalignment caused by pose errors and map the simultaneous object information in heterogeneous information to a unified coordinate system to obtain more accurate perception results. Vehicle and infrastructure sensors differ significantly in configurations, including types, noise levels, installation heights, etc. For instance, as shown in Figure 1, the data-level domain gap between the LiDAR point clouds of the vehicle and the roadside infrastructure is significant. These disparities in perception domains present unique challenges in designing collaborative fusion models.

{
\begin{figure}[t] 
\vspace{0.25cm}
    \centering
    \begin{subfigure}[b]{0.22\textwidth} 
        \centering
        \includegraphics[width=\textwidth]{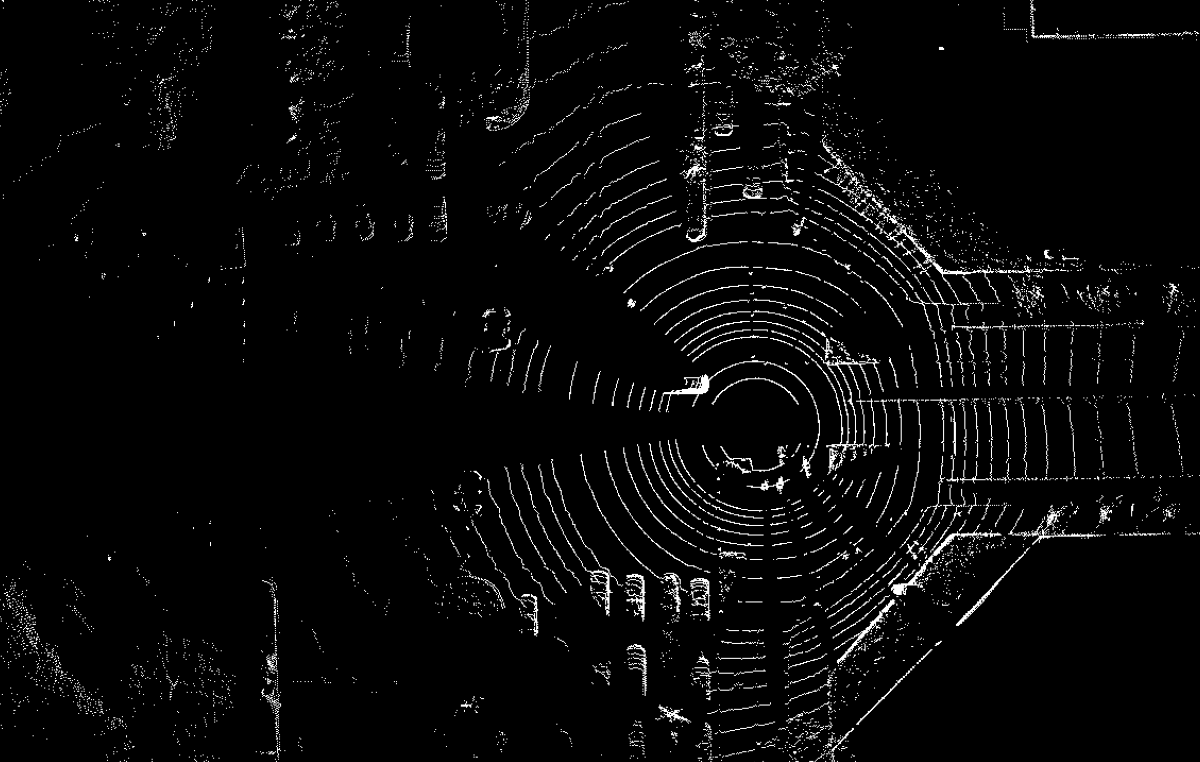} 
        \caption{Vehicle side}
        \label{fig:sub-a}
    \end{subfigure}
    \hspace{0.01\textwidth} %
    \begin{subfigure}[b]{0.22\textwidth}
        \centering
        \includegraphics[width=\textwidth]{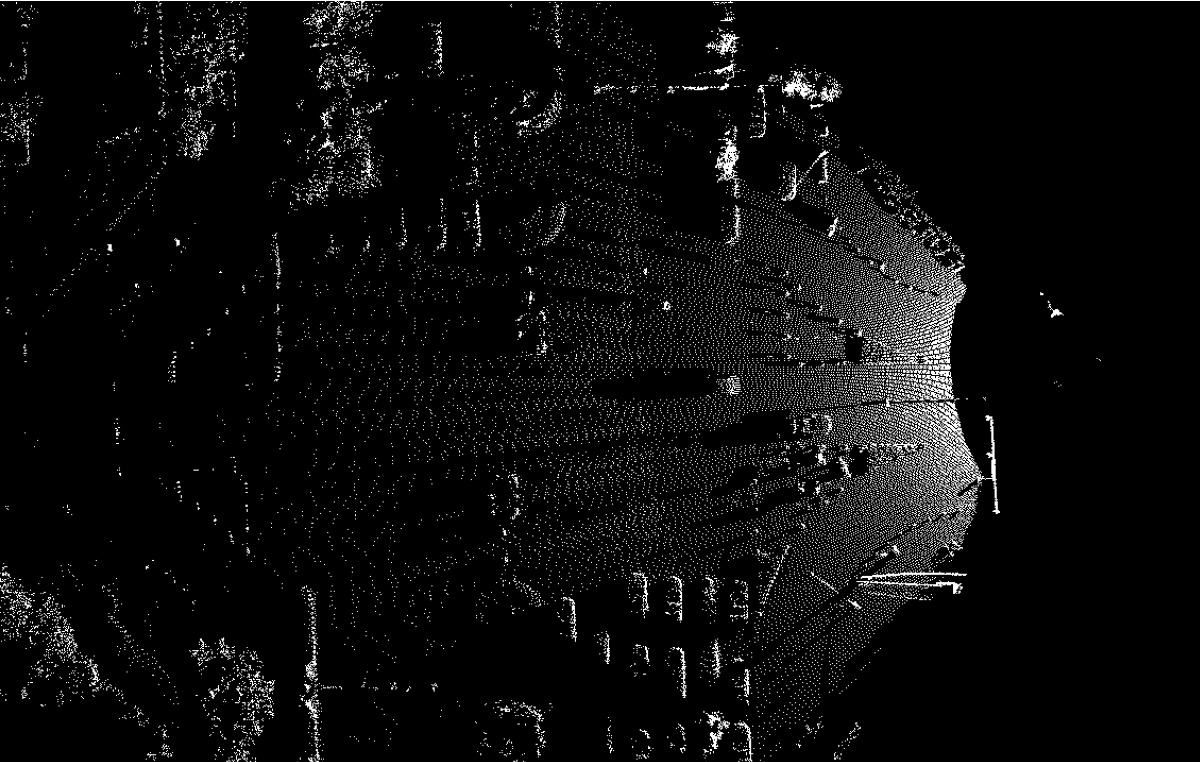} 
        \caption{Roadside infrastructure }
        \label{fig:sub-b}
    \end{subfigure}
    
    \caption{A sample from Dair-V2X illustrating the domain gap between vehicle (40-line LiDAR) and infrastructure (300-line LiDAR) point clouds .}
    \label{fig:main}
    
\end{figure}}

In this paper, we propose a V2X feature-level vehicle-infrastructure collaborative perception framework. This framework addresses two critical issues: domain gaps and pose errors. It employs a Knowledge Distillation Framework to learn domain-invariant representations from multi-source data. Additionally, a residual network-based Feature Compensation Module reduces the feature distribution gap between the vehicle and the infrastructure. Historical bird’s-eye-view (BEV) information is incorporated as supplementary input, enabling the model to capture the potentially important information in the historical frame and comprehensively understand the current scene. The Collaborative Fusion Module captures heterogeneous representations from the vehicle and the infrastructure through a heterogeneous multi-agent self-attention mechanism. By modeling the complex interactions between these agents, this module employs a refined mechanism of spatial information transmission and aggregation to overcome cross-domain perception challenges. In order to sample the feature offset caused by pose errors, we introduce a deformable attention module. By dynamically adjusting the positions of the sampling points, the most critical regions of input features are adaptively selected to focus on.

Extensive experiments conducted on the real-world DAIR-V2X dataset demonstrate that our proposed method significantly improves the performance of V2X LiDAR-based 3D object detection. Our proposed V2X-DGPE outperforms SOTA method DI-V2X\cite{div2xli2024di}  by 1.1\%/3.3\% for AP@0.5/0.7. Furthermore, under various pose noise levels of Gaussian and Laplace noise, V2X-DGPE achieves state-of-the-art performance. Our contributions are:

• We propose V2X-DGPE, a novel collaborative LiDAR-based 3D detection framework to address the challenges of domain gaps caused by heterogeneous perception nodes and unknown pose errors. This framework achieves both high detection accuracy and exceptional robustness.

• We integrate historical information into the framework and develop a Collaborative Fusion Module. This module leverages a heterogeneous self-attention mechanism and a deformable self-attention mechanism to effectively model heterogeneous interactions and enable adaptive sampling.

• Extensive experiments conducted on the real-world DAIR-V2X dataset demonstrate that V2X-DGPE effectively addresses domain gaps and unknown pose errors, achieving superior accuracy and robustness in 3D detection performance.





\section{RELATED WORK}

Collaborative perception fusion is a form of multi-source information fusion \cite{wang2019multi}. Compared to early fusion \cite{chen2019cooper} and late fusion \cite{35steinbaeck2018design,36wu2017lidar,zhao2017cooperative}, intermediate fusion \cite{liu2020who2com} strikes an effective balance between accuracy and transmission bandwidth. OPV2V \cite{xu2022opv2v} reconstructs a local graph for each vector in the feature graph, where feature vectors of the same spatial position of different vehicles are regarded as nodes and their mutual connections are regarded as edges of the local graph. F-Cooper \cite{chen2019f} proposes two fusion schemes based on point cloud features. The voxel feature fusion scheme directly fuses the features generated by the vehicle voxel feature encoding (VEF) to generate a spatial feature map. The spatial feature fusion scheme uses voxel features from VEF for individual vehicles to generate local spatial feature maps, which are then fused into an overall spatial feature map. V2VNet \cite{wang2020v2vnet} employs a convolutional GRU network to aggregate feature information shared by nearby vehicles, and uses a variational image compression algorithm to compress feature representations through multiple communication rounds. When2com \cite{liu2020when2com} introduces an asymmetric attention mechanism to select the most relevant communication partners and constructs a sparse communication graph. Where2comm \cite{hu2022where2comm} constructs a spatial confidence map for each agent, which informs the agent’s communication decisions regarding specific areas.

In the vehicle-infrastructure collaboration scenario, the difference in the perception domain of heterogeneous nodes presents significant challenges for collaborative information fusion. DiscoNet \cite{disconetli2021learning} applies knowledge distillation to multi-agent collaborative graph training, using the teacher model to guide the feature map generated by the student model after collaborative fusion. DI-V2X \cite{div2xli2024di} also adopts a Knowledge Distillation Framework to learn domain-invariant feature representations, reducing domain discrepancies. However, DI-V2X aligns student and teacher features before collaborative fusion, introducing unnecessary alignment that leads to the loss of original feature information. \cite{xu2023bridging} presents a Learnable Resizer and a sparse cross-domain transformer, employing adversarial training to bridge the domain gap. Heterogeneous graph transformer \cite{hu2020heterogeneous} excels at capturing the heterogeneity of multiple agents. Building on the inspiration from V2X-ViT \cite{v2xvitxu2022v2x}, we incorporated a heterogeneous multi-head self-attention module into the Collaborative Fusion Module.

Sensor asynchronous triggering, transmission delays, and noises contribute to agent pose errors. MASH \cite{mash2021overcoming} constructs similarity volumes and explicitly learns pixel correspondences to avoid incorporating noisy poses in inference. Extensions of Vision Transformer \cite{vitdosovitskiy2020image}, such as Swin \cite{liu2021swin}, CSwin \cite{dong2022cswin}, Twins \cite{chu2021twins}, and window \cite{winwang2022uformer} introduce window mechanisms into self-attention to capture both global and local interactions. V2X-ViT \cite{v2xvitxu2022v2x} further employs multi-scale window attention to integrate long-range information and local details to address pose errors. FPV-RCNN \cite{fpvrcnn2022keypoints} infers semantic labels of key points and corrects pose errors based on agents' correspondence. CoAlign \cite{coalignlu2023robust} proposes an agent-object pose graph that corrects relative poses among multiple agents by promoting the consistency of relative poses. However, the performance of these methods on real-world datasets still leaves much room for improvement.

\section{METHODOLOGY}
In collaborative perception, asynchronous triggering of sensors and communication transmission introduce time delays. Addressing feature misalignment due to time delays and pose errors is critical. This paper proposes a feature alignment method to achieve accurate spatial-temporal alignment, handle unknown pose errors, and map simultaneous information of heterogeneous sensing nodes to a unified coordinate system. Furthermore, significant variations in sensor configurations, such as type, noise levels, and installation heights, exacerbate the challenges. To tackle domain gaps among heterogeneous nodes, this paper designs collaborative fusion and object detection algorithms, leveraging the characteristics of intelligent agents for adaptive information integration.


{
\begin{figure*}
\begin{center}
\includegraphics[width=1\textwidth]{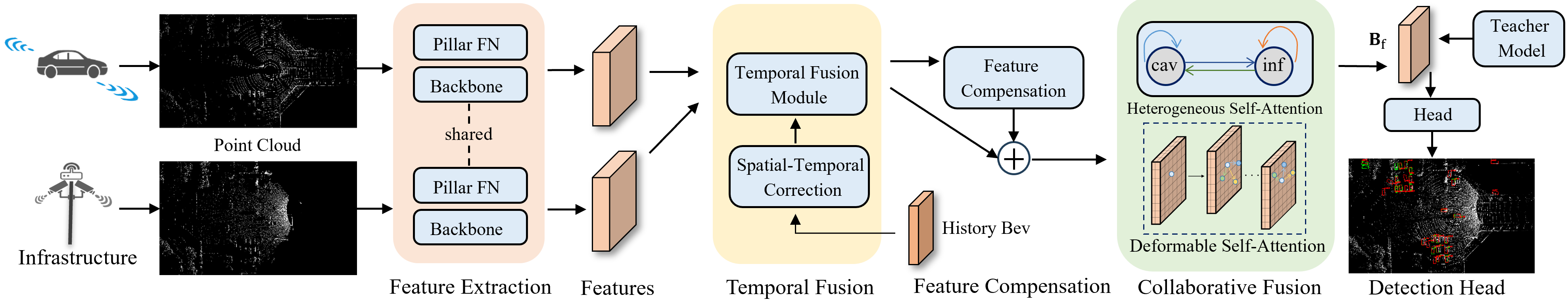}
   \caption{\textbf{Overview architecture of V2X-DGPE}. It employs a Knowledge Distillation Framework, comprising five key components arranged sequentially: BEV Feature Extraction Module, Temporal Fusion Module, Feature Compensation Module, Collaborative Fusion Module, and the Detection Head.}
   \label{fig:short}
\end{center}
\end{figure*}}

\subsection{Overall Architecture}

The overall architecture of the proposed framework is illustrated in Figure 2. The model utilizes a teacher-student Knowledge Distillation Framework. From left to right, the framework includes BEV Feature Extraction Module, Temporal Fusion Module, Feature Compensation Module, Collaborative Fusion Module, and the Detection Head. Guided by the teacher model, the student model learns domain-invariant representations of multi-source data in the vehicle-infrastructure collaborative scenario. The PointPillars method is employed to extract bird's-eye view (BEV) features. A detection head predicts object categories and regression results. The student model acquires  \( \mathbf{B}_{v} \) and  \( \mathbf{B}_{i} \) features using the PointPillars model. After passing through the Collaborative Fusion Module, the fusion feature  \( \mathbf{B}_{f} \) is generated. Alignment of the student feature \( \mathbf{B}_{f} \) with the teacher BEV feature \( \mathbf{B}_{t} \) occurs after the fusion module. Aligning the features before fusion would impose unnecessary constraints and distortions, resulting in losing original feature information.
 
After extracting features from both the vehicle and infrastructure point clouds, the BEV features from the infrastructure are sent to the vehicle and then input into the Temporal Fusion Module. Historical BEV is introduced to augment the current detection data. Following temporal fusion of the current and historical frames, they are passed into the Feature Compensation Module to reduce the feature distribution gap between the vehicle and infrastructure. The features are then processed through the Collaborative Fusion Module. After fusion, the student features are aligned with the teacher BEV features, and the final object detection results are obtained through the Detection Head.

\subsection{Feature Extraction Module}
First, the original point cloud is projected into a unified coordinate system and converted into pillars. Given the inference latency, the PointPillars method \cite{lang2019pointpillars} is employed as it avoids 3D convolutions, reduces latency, and is memory-efficient. After projection, the original point cloud is converted into a stacked columnar tensor. The tensor is then converted into a 2D pseudo image and input into Backbone to generate the BEV feature map. The BEV feature map \( \mathbf{B}_{v}^{t} \in \mathbb{R}^{H \times W \times C} \) represents the height (H), width (W), and channel (C) features of the ego-vehicle at time t.
\subsection{Temporal Fusion Module}


Affine transformation and resampling are employed to spatially correct the historical feature \cite{jaderberg2015spatial}. The process begins by utilizing the six-degree-of-freedom coordinates \(\boldsymbol{x}_{t-1} \) and \(\boldsymbol{x}_{t}\) of the ego-vehicle center at times t-1 and t, respectively, where each coordinate represents [x, y, z, roll, yaw, pitch]. These coordinates are used to compute the transformation matrices \(\mathbf{W}_{t-1}\) and \(\mathbf{W}_t\), which transform \(\boldsymbol{x}_{t-1} \) and \(\boldsymbol{x}_{t}\) into the world coordinate system. Next, the inverse transformation matrix \(\mathbf{W}_t^{-1}\) is derived to map the ego-vehicle coordinates at time t from the world coordinate system to the space rectangular coordinate system. By performing matrix multiplication of \(\mathbf{W}_t^{-1}\) and \(\mathbf{W}_{t-1}\), the direct transformation matrix \( \mathbf{T}\) is calculated,which maps \(\boldsymbol{x}_{t-1}\) directly to \(\boldsymbol{x}_{t}\). From this direct transformation matrix, a discretized affine transformation matrix \(\mathbf{T}_{affine}\) is generated. The affine matrix is expanded to a homogeneous matrix and adapted to the input and object sizes by normalization. Subsequently, the inverse transformation matrix is computed to generate the sampling grid \(\mathbf{G}\). Finally, the historical feature \(\mathbf{B}_{history}\) is resampled and transformed using bilinear interpolation based on the sampling grid, ensuring that the transformed historical features are spatially aligned with the current frame's features.

{
\begin{align*}
\mathbf{T} &= \mathbf{W}_t^{-1} \cdot\mathbf{W}_{t-1} \tag{1} \\[0.1cm]
\mathbf{T}_{affine}&=
\begin{bmatrix}
a_{11} & a_{12} & m_x \\
a_{21} & a_{22} & m_y
\end{bmatrix} \tag{2} \\[0.1cm]
\mathbf{T}_{norm} &= \text{Normalize}(\mathbf{T}_{affine},(H,W),(H^{'},W^{'})) \tag{3}\\[0.1cm]
\mathbf{G} &= \text{AffineGrid}(\mathbf{T}_{norm}^{-1},(H^{'},W^{'}))\tag{4} \\[0.1cm]
\mathbf{B}_{history} &= \text{GridSample}(\mathbf{B}_{t-1},\mathbf{G},\text{mode}=\text{'bilinear'}) \tag{5}
\end{align*}
}

Following spatial-temporal correction, the coordinates of the historical features align with the current feature center points. However, local features remain misaligned due to the movement of other vehicles during transmission. The known time delay is incorporated into the embedding representation. Time delay \(\Delta t_{i} \) and channel C are used as variables in a sine function for initialization, followed by input into the linear layer for projection \cite{hu2020heterogeneous}. This learnable projected embedding representation is directly added to the all detected objects features, enabling motion compensation.

 To effectively extract features within a relatively shallow structure, we design a Temporal Fusion Module based on a residual network, illustrated in Figure 3(a). The current feature \(\mathbf{B}_{current}\) and corrected historical feature \(\mathbf{B}_{history}\) from the vehicle or infrastructure are extracted as inputs. Those two BEV features are concatenated along the channel dimension, followed by feature extraction and dimensionality reduction through convolution operations. The second convolution layer further processes the features output from the first layer. The output of the second convolution layer is then added to the current moment's features via a residual connection, merging features from different moments while retaining the current moment's feature information. The Temporal Fusion Module is designed with a relatively simple and lightweight structure. Although the obtained feature \(\mathbf{B}_{temporal}\) incorporates historical information, it still needs to be input into the Collaborative Fusion Module for further feature extraction and fusion.

 {
\begin{figure}
\begin{center}
\includegraphics[width=0.49 \textwidth]{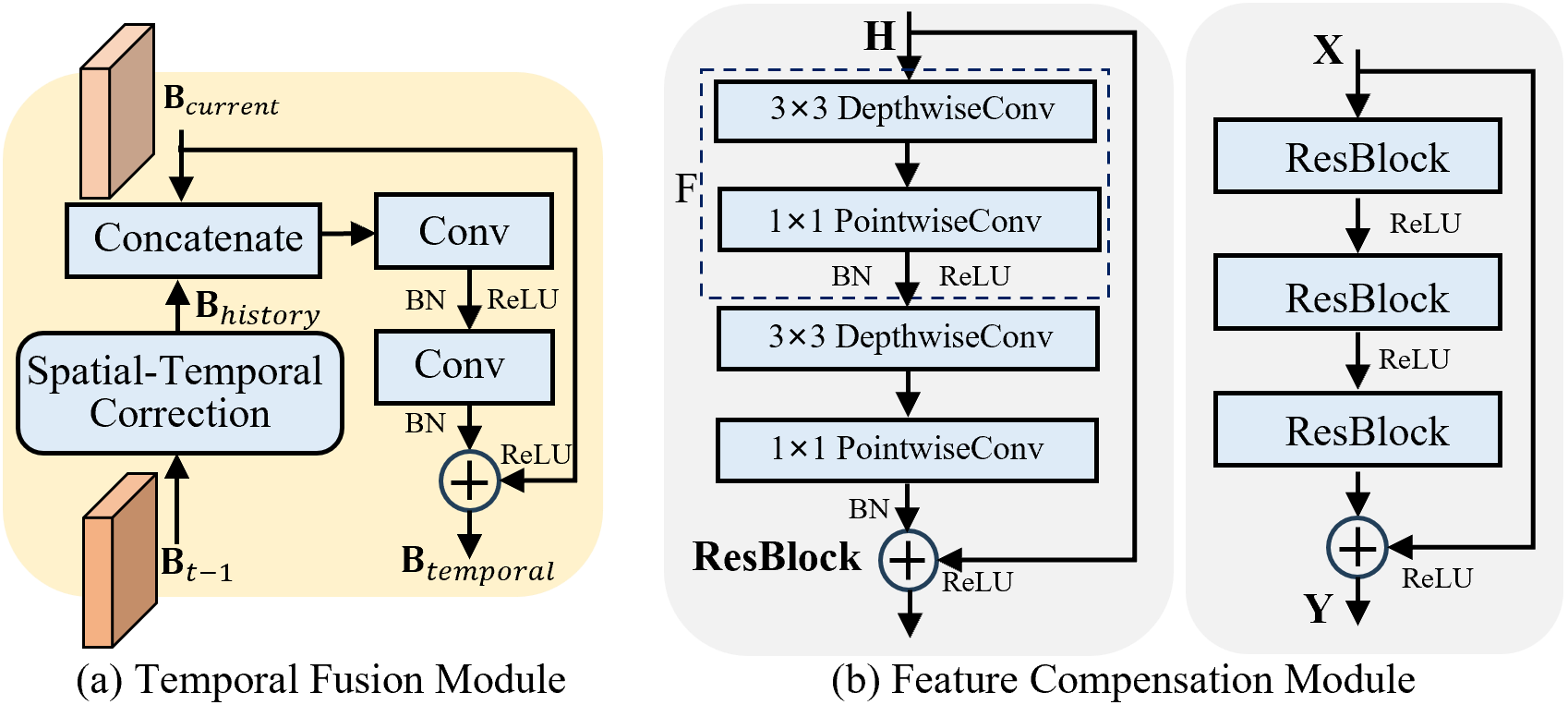}
   \caption{Illustration of the Temporal Fusion Module and Feature Compensation Module.}
   \label{fig:short}
\end{center}
\end{figure}
}

\subsection{Feature Compensation Module}
In order to reduce the distribution gap between the vehicle and infrastructure BEV features before collaborative fusion, we propose a feature compensation method, illustrated in Figure 3(b). The Feature Compensation Module operates on the original input features and generates the compensation feature through a series of residual blocks. Its structure is similar to the classic residual network, and the output of each residual block is regulated by specific weights. Three consecutive residual blocks are employed, with the contribution of each residual block modulated by adjustable weights. Each residual block cascades two depthwise separable convolutions, followed by residual connections to merge the input features with the convolved features. Depthwise separable convolutions are used to maintain computational efficiency. This convolution method consists of two stages: depthwise convolution and pointwise convolution. Compared to traditional convolution, depthwise separable convolution significantly reduces the number of parameters and computational complexity, while maintaining effective feature extraction capabilities. Finally, the feature compensation map is scaled and combined with the original input features to obtain the enhanced features. We employ KL divergence to quantify the distribution gap between the vehicle and infrastructure. The Feature Compensation Module enhances the network’s expressiveness through lightweight feature augmentation while preserving the input feature information.

{
\begin{flalign*}
&\text{F} = \text{DepthwiseSeparableConv}() \tag{6} \\[0.1cm]
&\text{ResBlock}_i(\mathbf{H}, \mathit{weight}_i)= \text{F}_2(\text{F}_1(\mathbf{H})) + \mathit{weight}_i \cdot \mathbf{H} \tag{7} \\[0.1cm]
&\mathbf{X}_{i+1} = \text{ResBlock}(\mathbf{X}_{i}, \mathit{weight}_i), \quad i \in [0, 2] \tag{8} \\[0.1cm]
&\mathbf{Y} = \text{ReLU}(\mathbf{X} + \mathit{weight} \cdot \mathbf{X}_2)& \tag{9} 
\end{flalign*}
}



{
\begin{figure*}
\begin{center}
\includegraphics[width=0.9\textwidth]{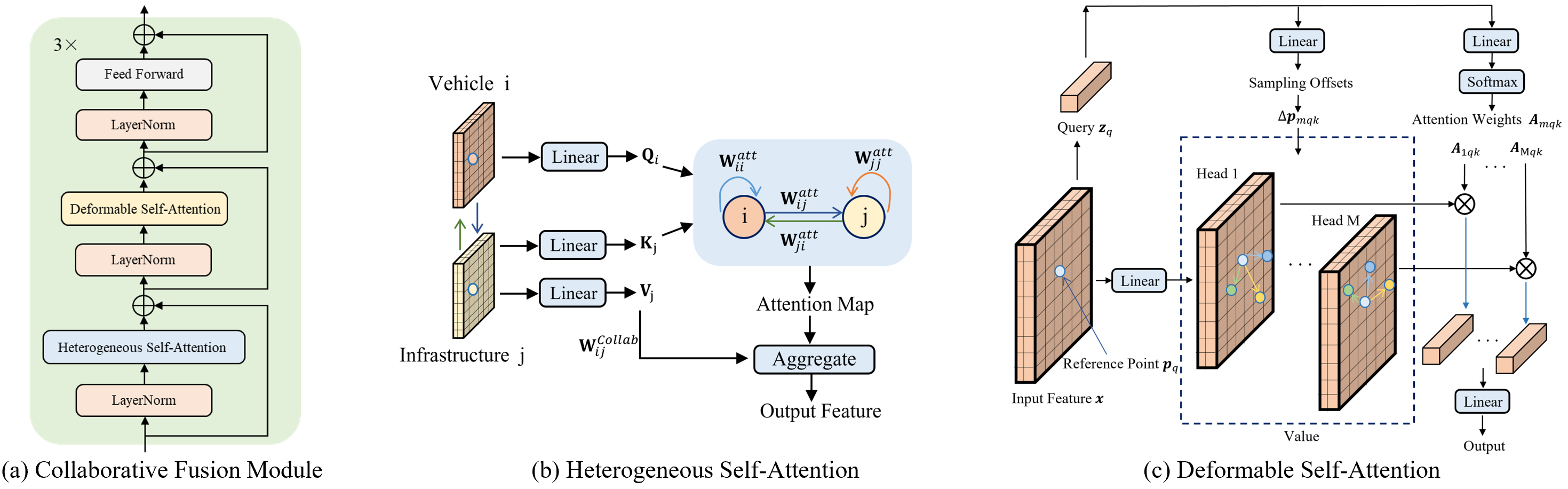}
   \caption{(a) The architecture of the Collaborative Fusion Module. (b) Illustration of the heterogeneous self-attention module. (c) Illustration of the deformable self-attention module.}
   \label{fig:short}
\end{center}
\end{figure*}
}
\subsection{Collaborative Fusion Module}
The Collaborative Fusion Module, illustrated in Figure 4, is composed of two components: heterogeneous self-attention and deformable self-attention. Depending on the type of agent (ego-vehicle or infrastructure), the heterogeneous multi-agent self-attention Module \cite{v2xvitxu2022v2x} applies a specific linear projection to each agent's features, projecting them onto the query (\(\mathbf{Q})\), key (\(\mathbf{K}\)), and value (\(\mathbf{V}\)) matrices, as shown in the formula below. The attention and value matrix weights are computed based on the agent type. For each agent pair \(i\) and \(j\), the attention and value matrix weights are determined by learnable relationship-specific parameters. Different types of agent combinations, including vehicle-to-vehicle, infrastructure-to-infrastructure, vehicle-to-infrastructure, and infrastructure-to-vehicle, have distinct weights. Interactions between different agent types are captured through learnable relationship-specific weights.
{
\begin{align*}
\mathbf{Q}_i^h &= \mathbf{X}_i \mathbf{W}_{qi}^h \tag{10} \\
\mathbf{K}_j^h &= \mathbf{X}_j \mathbf{W}_{kj}^h \tag{11} \\
\mathbf{V}_j^h &= \mathbf{X}_j \mathbf{W}_{vj}^h \tag{12}
\end{align*}
}

The calculation formula for the attention map of the attention graph is provided below. \(\mathbf{W}^{att}_{ij}\) denotes the attention weight based on the type relationship. If \(i=j\), it indicates self-attention, including \(\mathbf{W}^{att}_{ii}\), \(\mathbf{W}^{att}_{ij}\), \(\mathbf{W}^{att}_{ji}\), \(\mathbf{W}^{att}_{jj}\). The attention graph calculates the attention scores of each agent with all other agents, including vehicle self-attention, infrastructure self-attention, and heterogeneous attention between vehicle and infrastructure. The multi-head attention mechanism is applied when obtaining the attention matrix \(\textbf{AttentionMap}\) and the value matrix \(\mathbf{V}^{Collab}\). The formula is as follows:

{
\begin{equation}
\textbf{AttentionMap}_{ij}^h = \text{Softmax} \left( \frac{\mathbf{Q}_i^h \cdot \mathbf{W}^{att,h}_{ij} \cdot \left( \mathbf{K}_j^h \right)^T}{\sqrt{d_k}} \right) \tag{13} 
\end{equation}
}

The value matrix weight \(\mathbf{W}^{Collab}_{ij}\) weights the value matrix to represent the collaborative interaction between the vehicle and infrastructure. This process represents how the agent integrates features from both itself and other agents.
{
\begin{equation}
\mathbf{V}^{Collab,h}_{ij} = \mathbf{W}^{Collab,h}_{ij} \cdot \mathbf{V}_j^h \tag{14}
\end{equation}
}

Finally, the obtained attention matrix \(\textbf{AttentionMap}_{ij}\) is used to perform a weighted summation on the value matrix \( \mathbf{V}^{Collab}_{ij}\), producing the collaborative fusion feature output. Multiple heads in the multi-head attention mechanism are aggregated, with the output of all heads \(h\) calculated and concatenated to form the collaborative fusion feature \(\mathbf{B}^{Fusion}\). The formula is as follows, where \(j\) traverses all agents, including both itself and heterogeneous agents:
{
\begin{align}
\mathbf{B}_{i}^{h} &= \sum_{j} \left( \textbf{AttentionMap}_{ij}^h \cdot \mathbf{V}^{Collab,h}_{ij} \right) \tag{15} \\
\mathbf{B}^{\textup{Fusion}} &= \text{Concat}(\mathbf{B}_i^1, \mathbf{B}_i^2, \dots, \mathbf{B}_i^H) \cdot \mathbf{W}^B \tag{16} 
\end{align}
}


To address the feature shift caused by time delays or pose errors, we introduce a deformable attention module in the Collaborative Fusion Module. Unlike traditional multi-head self-attention mechanisms, the deformable attention mechanism \cite{zhu2020deformable} focuses on a subset of key positions. These positions are not fixed; instead, they are dynamically predicted by the model as sampling points. The model learns several sets of query-agnostic offsets to shift keys and values toward important areas. Specifically, for each attention module, reference points are first generated as a uniform grid over the input data. The offset network then takes the query features as input and generates corresponding offsets for all reference points. For each reference point, the model predicts multiple offsets, which are added to the reference point’s position to define the final sampling point positions. The model extracts features from the corresponding feature maps at these dynamic sampling point positions and performs a weighted summation of these features to obtain the final output. This module can be viewed as a spatial adaptive mechanism. The model adaptively selects the most relevant parts of input by dynamically adjusting the sampling points. This approach is particularly effective in handling pose errors. Additionally, deformable attention focuses only on a small number of sampling points, significantly reducing computational costs. We use the self-attention mechanism , where both the query \(\boldsymbol{z}_q\) and value \(\boldsymbol{x}\) matrices are derived from the input features. The formula is as follows:


{
\begin{equation}\label{eqn:1}
\begin{array}{l}
\text{DeformAttn}( \boldsymbol{z}_q, \boldsymbol{p}_q, \boldsymbol {x}) =\tag{17}  \\[10pt] 
\sum_{m=1}^M \mathbf{W}_m \left[ \sum_{k=1}^K \mathbf{A}_{mqk} \cdot \mathbf{W}_m' \boldsymbol {x}(\boldsymbol{p}_q + \Delta \boldsymbol{p}_{mqk}) \right]
\end{array}
\end{equation}
}
{
\begin{table}[!t]

\label{tab:fonts}
\begin{center}    
\small
\begin{tabular}{p{0.27\linewidth}|p{0.24\linewidth}|c|c}
\hline
Models & Publication & AP@0.5 &AP@0.7 \\ \hline
 No Fusion \cite{lang2019pointpillars} & CVPR 2019 & 0.652 & 0.543 \\ 
 Late Fusion \cite{lang2019pointpillars}& CVPR 2019 & 0.637 & 0.453 \\ 
 Early Fusion \cite{lang2019pointpillars}& CVPR 2019 & 0.766 & 0.641 \\ \hdashline
  PP-IF \cite{lang2019pointpillars}& CVPR 2019 & 0.725 & 0.544 \\ 
 V2VNet \cite{wang2020v2vnet} & ECCV 2020 & 0.708 & 0.476 \\ 
 DiscoNet \cite{disconetli2021learning} & NeurIPS 2021 & 0.739 & 0.596 \\ 
  OPV2V \cite{xu2022opv2v} & ICRA 2022 & 0.733 & 0.553 \\ 
  V2X-ViT \cite{v2xvitxu2022v2x} & ECCV 2022 & 0.718 & 0.549 \\ 
 Where2comm\cite{hu2022where2comm}  & NeurIPS 2022 & 0.741 & 0.597 \\ 
 CoAlign \cite{coalignlu2023robust}  & ICRA 2023 & 0.746 & 0.604 \\ 
 DI-V2X \cite{div2xli2024di}  & AAAI 2024 & 0.788 & 0.662 \\ \hline
  Ours & & \textbf{0.797} & \textbf{0.684} \\ \hline
\end{tabular}

\end{center}
\caption{\textbf{Detection performance comparison on DAIR-V2X dataset}. We compared the detection performance of various state-of-the-art models using the BEV AP@0.5 and AP@0.7 metrics. BEV AP@0.5 represents the Average Precision (AP) for 3D object detection in the Bird's-Eye View (BEV) at IoU=0.5.
 } 
\end{table}
}

{
\begin{table*}[!t]
\centering
\small
\begin{tabular}{l|l|cccc|cccc}
\hline
\multicolumn{2}{c|}{Method/Metric} & \multicolumn{4}{c|}{AP@0.5} & \multicolumn{4}{c}{AP@0.7} \\ 
\hline
\multicolumn{2}{c|}{Noise Level ${\sigma_t}/{\sigma_r}$ (m/°)}& 0.0/0.0 & 0.2/0.2 & 0.4/0.4 & 0.6/0.6 & 0.0/0.0 & 0.2/0.2 & 0.4/0.4 & 0.6/0.6 \\
\hline
\multirow{4}{*}{\shortstack[l]{w/o\\robust\\design}} & F-Cooper \cite{chen2019f} & 0.734 & 0.723 & 0.705 & 0.692 & 0.559 & 0.552 & 0.542 & 0.538 \\
 & V2VNet \cite{wang2020v2vnet} & 0.664 & 0.649 & 0.623 & 0.599 & 0.402 & 0.388 & 0.367 & 0.350 \\
 & DiscoNet \cite{disconetli2021learning} & 0.736 & 0.726 & 0.708 & 0.697 & 0.583 & 0.576 & 0.569 & 0.566 \\
 & \(\text{OPV2V}_{\text{pointpillar}}\) \cite{xu2022opv2v} & 0.733 & 0.723 & 0.708 & 0.697 & 0.553 & 0.547 & 0.540 & 0.538 \\
\hline
\multirow{7}{*}{\shortstack[l]{w/\\robust\\design}} & MASH \cite{mash2021overcoming} & 0.400 & 0.400 & 0.400 & 0.400 & 0.244 & 0.244 & 0.244 & 0.244 \\
 & FPV-RCNN \cite{fpvrcnn2022keypoints} & 0.655 & 0.631 & 0.580 & 0.581 & 0.505 & 0.459 & 0.420 & 0.410 \\
 & \(\text{V2VNet}_{\text{robust}}\) \cite{v2vnetrobust2021learning}& 0.660 & 0.655 & 0.646 & 0.636 & 0.486 & 0.483 & 0.478 & 0.475 \\
 & V2X-ViT \cite{v2xvitxu2022v2x}& 0.704 & 0.700 & 0.700 & 0.694 & 0.531 & 0.529 & 0.525 & 0.522 \\
 & Coalign \cite{coalignlu2023robust}& 0.746 & 0.738 & 0.720 & 0.700 & 0.604 & 0.588 & 0.579 & 0.570 \\
 & DI-V2X \cite{div2xli2024di}& 0.787 & 0.765 & 0.720 & 0.688 & 0.658 & 0.615 & 0.588 & 0.577 \\ \cline{2-10}
 & Ours & \textbf{0.790} & \textbf{0.771} & \textbf{0.727} & \textbf{0.701} & \textbf{0.670} & \textbf{0.629} & \textbf{0.601} & \textbf{0.594} \\
\hline
\end{tabular}
\caption{\textbf{Detection performance comparison on DAIR-V2X dataset of methods with and without robust design under Gaussian pose noises}. All models are trained with pose noise, where \( \sigma_t = 0.2  \)\text{m} and \( \sigma_r = 0.2^\circ \), following a Gaussian distribution. The models are evaluated at various Gaussian noise levels. The results demonstrate that V2X-DGPE significantly outperforms existing methods across various noise levels, showcasing superior robustness to pose errors.
}
\label{tab:methods_comparison}
\end{table*}}
{
\begin{table}
    \setlength{\tabcolsep}{1pt}
    \label{tab:fusion_comparison}
    \begin{center}
    \begin{tabular}{lccccccc}
    
    \hline
    \rule{0pt}{5ex} \footnotesize \text{} &\footnotesize \shortstack{\text{Knowledge} \\ \text{Distillation}}& \footnotesize \shortstack{\text{Feature} \\ \text{Compensation}} &\footnotesize \shortstack{\text{Collaborative} \\ \text{Fusion}} & \footnotesize \shortstack{\text{Temporal} \\ \text{Fusion}} &\footnotesize  \text{AP@0.5} &\footnotesize \text{AP@0.7} \\ \hline
    \footnotesize \text{PP-IF} & & & & &\footnotesize 0.725 & \footnotesize 0.544 \\
    &\footnotesize \checkmark & & & & \footnotesize 0.746 & \footnotesize 0.561 \\
    &\footnotesize \checkmark & \footnotesize \checkmark & & &\footnotesize 0.751 & \footnotesize 0.574 \\
    &\footnotesize \checkmark & \footnotesize \checkmark & \footnotesize \checkmark & & \footnotesize 0.783 &\footnotesize 0.662 \\
    &\footnotesize \checkmark & \footnotesize \checkmark & \footnotesize \checkmark & \footnotesize \checkmark & \footnotesize 0.797 & \footnotesize 0.684 \\ \hline
    \end{tabular}
    \caption{\textbf{Ablation studies}. PP-IF refers to intermediate fusion based on PointPillars \cite{lang2019pointpillars}.}
    \end{center}
\end{table}
}
\section{EXPERIMENTS}
\subsection{DAIR-V2X}
Most existing algorithms have been tested on simulated datasets and have yet to be validated in real-world scenarios. We employed the real-world V2X collaborative perception dataset DAIR-V2X \cite{yu2022dair} for evaluation. A well-equipped vehicle is deployed through the intersection in the data collection area, with separate recordings of vehicle frames and infrastructure frames. The dataset comprises 100 manually selected scenes of 20-second vehicle passages through the intersection, yielding 9,000 synchronized frame pairs sampled at 10Hz. The vehicle is equipped with a 40-line LiDAR, providing a \( 360^\circ \) horizontal field of view (FOV). The roadside infrastructure is equipped with a 300-line LiDAR with a horizontal field of view (FOV) of \( 100^\circ \).
\subsection{Experimental Setup}
\textbf{Evaluation metrics}. Detection performance is evaluated using Average Precision (AP) at Intersection-over-Union (IoU) thresholds of 0.5 and 0.7.

\textbf{Implementation Details}. We set the point cloud range to \([-100, 100] \times [-40, 40] \times [-3.5, 1.5]\) meters in the vehicle coordinate system. For the PointPillar backbone, the voxel resolution in both height and width is set to 0.4m. We employed the Adam optimizer with an initial learning rate of 0.001, which decayes steadily by a factor of 0.1 at epochs 15, 30, and 40. The batch size is set to 4, and the model is trained for 45 epochs on a single NVIDIA A100.

\subsection{Performance Comparison}
We compare late fusion, early fusion, and intermediate fusion methods. The comparison results of various methods are shown in Table 1. Late fusion performs object detection on the vehicle or infrastructure separately, and then matches the object detection boxes to generate the final results. Early fusion directly transmits the original LiDAR point cloud, which is then fused after coordinate and time alignment. For intermediate fusion, we compare performance against the most advanced V2X collaborative perception methods. Specifically, Our proposed V2X-DGPE outperforms SOTA intermediate fusion methods DI-V2X\cite{div2xli2024di}  by 1.1\%/3.3\% for AP@0.5/0.7, significantly outperforming other intermediate fusion methods. This result demonstrates that our method can more efficiently model and leverage perception information across heterogeneous agents, leading to more accurate object detection.



{

\begin{table*}[!t]
\centering
\small
\begin{tabular}{l|l|cccc|cccc}
\hline
\multicolumn{2}{c|}{Method/Metric} & \multicolumn{4}{c|}{AP@0.5} & \multicolumn{4}{c}{AP@0.7} \\ 
\hline
\multicolumn{2}{c|}{Noise Level ${b_t}/{b_r}$ (m/°)}& 0.0/0.0 & 0.2/0.2 & 0.4/0.4 & 0.6/0.6 & 0.0/0.0 & 0.2/0.2 & 0.4/0.4 & 0.6/0.6 \\
\hline
\multirow{4}{*}{\shortstack[l]{w/o\\robust\\design}}  & F-Cooper \cite{chen2019f}& 0.734 & 0.716 & 0.694 & 0.681 & 0.559 & 0.552 & 0.541 & 0.530 \\
 & V2VNet \cite{wang2020v2vnet}& 0.665 & 0.639 & 0.606 & 0.585 & 0.401 & 0.379 & 0.356 & 0.342 \\
 & DiscoNet \cite{disconetli2021learning}& 0.736 & 0.718 & 0.700 & 0.688 & 0.583 & 0.574 & 0.567 & 0.563 \\
 &\(\text{OPV2V}_{\text{pointpillar}}\) \cite{xu2022opv2v}  & 0.733 & 0.718 & 0.701 & 0.692 & 0.553 & 0.546 & 0.538 & 0.536 \\
\hline
\multirow{7}{*}{\shortstack[l]{w/\\robust\\design}} & MASH \cite{mash2021overcoming} & 0.400 & 0.400 & 0.400 & 0.400 & 0.244 & 0.244 & 0.244 & 0.244 \\
 & FPV-RCNN \cite{fpvrcnn2022keypoints}& 0.654 & 0.609 & 0.564 & 0.553 & 0.504 & 0.436 & 0.417 & 0.431 \\
 & \(\text{V2VNet}_{\text{robust}}\) \cite{v2vnetrobust2021learning}  & 0.660 & 0.653 & 0.640 & 0.632 & 0.486 & 0.481 & 0.476 & 0.469 \\
 & V2X-ViT \cite{v2xvitxu2022v2x}& 0.705 & 0.695 & 0.680 & 0.667 & 0.531 & 0.527 & 0.521 & 0.517 \\
 & Coalign \cite{coalignlu2023robust} & 0.746 & 0.733 & 0.707 & 0.689 & 0.604 & 0.585 & 0.573 & 0.566 \\
 & DI-V2X \cite{div2xli2024di}& 0.787 & 0.750 & 0.701 & 0.677 & 0.659 & 0.607 & 0.578 & 0.569 \\ \cline{2-10}
 & Ours & \textbf{0.790} & \textbf{0.753} & \textbf{0.707} & \textbf{0.691} & \textbf{0.670} & \textbf{0.620} & \textbf{0.595} & \textbf{0.588} \\
\hline
\end{tabular}
\caption{\textbf{Detection performance comparison on DAIR-V2X dataset of methods with and without robust design under Laplace pose noises}. All models are trained with pose noise, where \( \sigma_t = 0.2  \)\text{m} and \( \sigma_r = 0.2^\circ \), following a Gaussian distribution. The models are evaluated at various noise levels, following a Laplace distribution. The results consistently outperform existing methods across all noise levels, confirming that V2X-DGPE is resilient to unexpected noises. 
}
\label{tab:methods_comparison}
\end{table*}}

{
\begin{figure*}
\begin{center}
\includegraphics[width=0.95\textwidth]{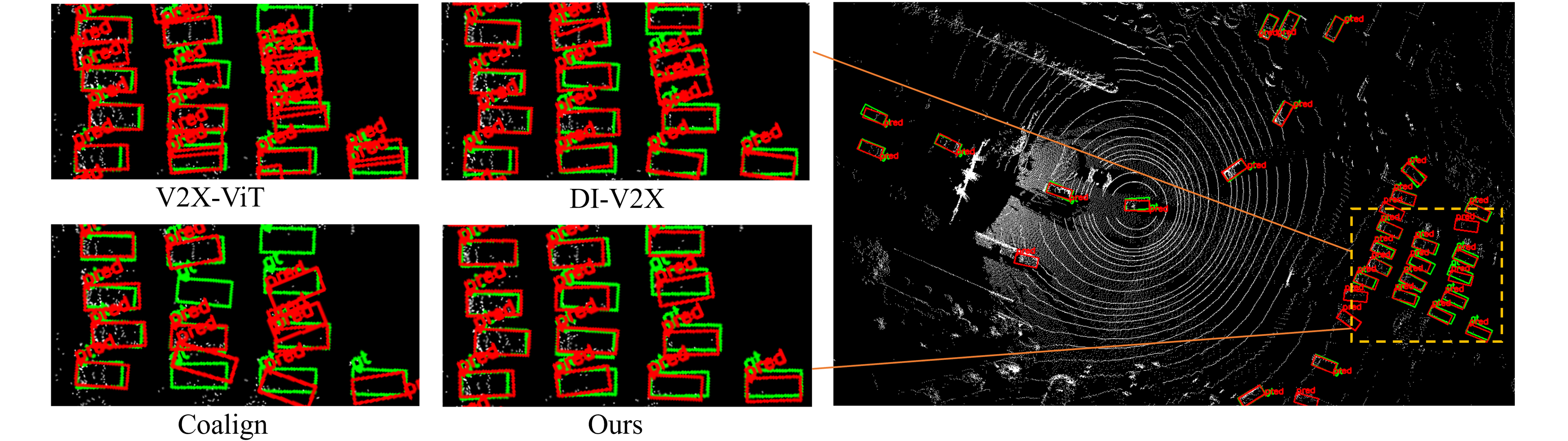}
   \caption{\textbf{Detection visualization of V2X-ViT, DI-V2X, Coalign, and V2X-DGPE under Gaussian noise} with \( \sigma_t = 0.4  \)\text{m} and \( \sigma_r = 0.4^\circ \). The green boxes represent the ground truth, while the red boxes represent the detected results. Compared to other advanced models, the proposed model, V2X-DGPE, demonstrates superior detection accuracy, with its detection boxes being noticeably more precise.}
   \label{fig:short}
\end{center}
\end{figure*}
}
\subsection{Ablation Studies}
As shown in Table 3, our experimental results indicate that the introduction of all modules significantly contributes to overall performance improvement. Specifically, the Collaborative Fusion Module improves AP@0.5 by 4.3\% and AP@0.7 by 15.3\%, while the Temporal Fusion Module further improves AP@0.5 by 1.8\% and AP@0.7 by 3.3\%, building on the Collaborative Fusion Module.
The Collaborative Fusion Module effectively captures heterogeneous representations from both vehicle and infrastructure, modeling complex interaction relationships. This module effectively addresses the domain gaps between data sources and enhances cross-domain perception accuracy through advanced spatial information transmission and aggregation mechanisms.

The Temporal Fusion Module further enhances the model's detection accuracy. This module captures potential critical information from the historical frame by efficiently fusing historical features with the current moment's features. This information supplements the model's input, enabling a more comprehensive understanding of the current scene based on continuous frame information, while mitigating information loss or false detection. 

In the knowledge distillation framework, the student model learns domain-invariant representations under the guidance of the teacher model. The Feature Compensation Module reduces the domain gaps of BEV features between the vehicle and infrastructure before collaborative fusion. The combined application of these modules significantly improves perception accuracy, particularly under the stringent standard of AP@0.7.

\subsection{Pose Errors Reslut}
To evaluate the 3D detection performance under pose errors, we compare V2X-DGPE against several existing approaches, both with and without pose-robust design. All models are trained with pose noise, where \( \sigma_t = 0.2  \)\text{m} and \( \sigma_r = 0.2^\circ \). The 2D global center coordinates \( x \) and \( y \) are perturbed with \( \mathcal{N}(0, \sigma_t) \) Gaussian noise, while the yaw angle \( \theta \) are perturbed with \( \mathcal{N}(0, \sigma_r) \) Gaussian noise. Pose noise during testing also follows the Gaussian distribution. We evaluate the model at various noise levels. As shown in Table 2, the results demonstrate that our model significantly outperforms existing methods, exhibiting superior robustness to pose errors.

Additionally, We train all models under Gaussian noise with \( \sigma_t = 0.2  \)\text{m} and \( \sigma_r = 0.2^\circ \), and test them under Laplace noise. As shown in Table 4, the test results consistently outperform existing methods across all noise levels, confirming that our model is resilient to unexpected noise. Our approach effectively mitigates pose errors, enhancing both detection accuracy and overall robustness.
\section{CONCLUSION}
In this paper, we propose a novel framework, V2X-DGPE, for robust collaborative perception. V2X-DGPE effectively reduces domain gaps between heterogeneous nodes and achieves high accuracy. V2X-DGPE outperforms SOTA intermediate fusion methods DI-V2X\cite{div2xli2024di} by 1.1\%/3.3\% for AP@0.5/0.7. Furthermore, V2X-DGPE demonstrates strong robustness in handling pose error noises. Under various pose noise levels of Gaussian and Laplace noise, V2X-DGPE achieves state-of-the-art performance. In future works, we will leverage more historical data and extend the integration of multimodal sensors to enhance V2X collaborative perception and prediction.

{\footnotesize
\bibliographystyle{ieee}
\bibliography{egbib}

\begin{thebibliography}{10}\itemsep=-1pt

\bibitem{chen2019f}
Q.~Chen, X.~Ma, S.~Tang, J.~Guo, Q.~Yang, and S.~Fu.
\newblock F-cooper: Feature based cooperative perception for autonomous vehicle edge computing system using 3d point clouds.
\newblock In {\em Proceedings of the 4th ACM/IEEE Symposium on Edge Computing}, pages 88--100, 2019.

\bibitem{chen2019cooper}
Q.~Chen, S.~Tang, Q.~Yang, and S.~Fu.
\newblock Cooper: Cooperative perception for connected autonomous vehicles based on 3d point clouds.
\newblock In {\em 2019 IEEE 39th International Conference on Distributed Computing Systems (ICDCS)}, pages 514--524. IEEE, 2019.

\bibitem{chu2021twins}
X.~Chu, Z.~Tian, Y.~Wang, B.~Zhang, H.~Ren, X.~Wei, H.~Xia, and C.~Shen.
\newblock Twins: Revisiting the design of spatial attention in vision transformers.
\newblock {\em Advances in neural information processing systems}, 34:9355--9366, 2021.

\bibitem{dong2022cswin}
X.~Dong, J.~Bao, D.~Chen, W.~Zhang, N.~Yu, L.~Yuan, D.~Chen, and B.~Guo.
\newblock Cswin transformer: A general vision transformer backbone with cross-shaped windows.
\newblock In {\em Proceedings of the IEEE/CVF conference on computer vision and pattern recognition}, pages 12124--12134, 2022.

\bibitem{vitdosovitskiy2020image}
A.~Dosovitskiy.
\newblock An image is worth 16x16 words: Transformers for image recognition at scale.
\newblock {\em arXiv preprint arXiv:2010.11929}, 2020.

\bibitem{mash2021overcoming}
N.~Glaser, Y.-C. Liu, J.~Tian, and Z.~Kira.
\newblock Overcoming obstructions via bandwidth-limited multi-agent spatial handshaking.
\newblock In {\em 2021 IEEE/RSJ International Conference on Intelligent Robots and Systems (IROS)}, pages 2406--2413. IEEE, 2021.

\bibitem{hu2022where2comm}
Y.~Hu, S.~Fang, Z.~Lei, Y.~Zhong, and S.~Chen.
\newblock Where2comm: Communication-efficient collaborative perception via spatial confidence maps.
\newblock {\em Advances in neural information processing systems}, 35:4874--4886, 2022.

\bibitem{hu2020heterogeneous}
Z.~Hu, Y.~Dong, K.~Wang, and Y.~Sun.
\newblock Heterogeneous graph transformer.
\newblock In {\em Proceedings of the web conference 2020}, pages 2704--2710, 2020.

\bibitem{jaderberg2015spatial}
M.~Jaderberg, K.~Simonyan, A.~Zisserman, et~al.
\newblock Spatial transformer networks.
\newblock {\em Advances in neural information processing systems}, 28, 2015.

\bibitem{lang2019pointpillars}
A.~H. Lang, S.~Vora, H.~Caesar, L.~Zhou, J.~Yang, and O.~Beijbom.
\newblock Pointpillars: Fast encoders for object detection from point clouds.
\newblock In {\em Proceedings of the IEEE/CVF conference on computer vision and pattern recognition}, pages 12697--12705, 2019.

\bibitem{lei2022latency}
Z.~Lei, S.~Ren, Y.~Hu, W.~Zhang, and S.~Chen.
\newblock Latency-aware collaborative perception.
\newblock In {\em European Conference on Computer Vision}, pages 316--332. Springer, 2022.

\bibitem{div2xli2024di}
X.~Li, J.~Yin, W.~Li, C.~Xu, R.~Yang, and J.~Shen.
\newblock Di-v2x: Learning domain-invariant representation for vehicle-infrastructure collaborative 3d object detection.
\newblock In {\em Proceedings of the AAAI Conference on Artificial Intelligence}, volume~38, pages 3208--3215, 2024.

\bibitem{li2022v2xsim}
Y.~Li, D.~Ma, Z.~An, Z.~Wang, Y.~Zhong, S.~Chen, and C.~Feng.
\newblock V2x-sim: Multi-agent collaborative perception dataset and benchmark for autonomous driving.
\newblock {\em IEEE Robotics and Automation Letters}, 7(4):10914--10921, 2022.

\bibitem{disconetli2021learning}
Y.~Li, S.~Ren, P.~Wu, S.~Chen, C.~Feng, and W.~Zhang.
\newblock Learning distilled collaboration graph for multi-agent perception.
\newblock {\em Advances in Neural Information Processing Systems}, 34:29541--29552, 2021.

\bibitem{liu2020when2com}
Y.-C. Liu, J.~Tian, N.~Glaser, and Z.~Kira.
\newblock When2com: Multi-agent perception via communication graph grouping.
\newblock In {\em Proceedings of the IEEE/CVF Conference on computer vision and pattern recognition}, pages 4106--4115, 2020.

\bibitem{liu2021swin}
Z.~Liu, Y.~Lin, Y.~Cao, H.~Hu, Y.~Wei, Z.~Zhang, S.~Lin, and B.~Guo.
\newblock Swin transformer: Hierarchical vision transformer using shifted windows.
\newblock In {\em Proceedings of the IEEE/CVF international conference on computer vision}, pages 10012--10022, 2021.

\bibitem{coalignlu2023robust}
Y.~Lu, Q.~Li, B.~Liu, M.~Dianati, C.~Feng, S.~Chen, and Y.~Wang.
\newblock Robust collaborative 3d object detection in presence of pose errors.
\newblock In {\em 2023 IEEE International Conference on Robotics and Automation (ICRA)}, pages 4812--4818. IEEE, 2023.

\bibitem{35steinbaeck2018design}
J.~Steinbaeck, C.~Steger, G.~Holweg, and N.~Druml.
\newblock Design of a low-level radar and time-of-flight sensor fusion framework.
\newblock In {\em 2018 21st Euromicro Conference on Digital System Design (DSD)}, pages 268--275. IEEE, 2018.

\bibitem{v2vnetrobust2021learning}
N.~Vadivelu, M.~Ren, J.~Tu, J.~Wang, and R.~Urtasun.
\newblock Learning to communicate and correct pose errors.
\newblock In {\em Conference on Robot Learning}, pages 1195--1210. PMLR, 2021.

\bibitem{wang2020v2vnet}
T.-H. Wang, S.~Manivasagam, M.~Liang, B.~Yang, W.~Zeng, and R.~Urtasun.
\newblock V2vnet: Vehicle-to-vehicle communication for joint perception and prediction.
\newblock In {\em Computer Vision--ECCV 2020: 16th European Conference, Glasgow, UK, August 23--28, 2020, Proceedings, Part II 16}, pages 605--621. Springer, 2020.

\bibitem{winwang2022uformer}
Z.~Wang, X.~Cun, J.~Bao, W.~Zhou, J.~Liu, and H.~Li.
\newblock Uformer: A general u-shaped transformer for image restoration.
\newblock In {\em Proceedings of the IEEE/CVF conference on computer vision and pattern recognition}, pages 17683--17693, 2022.

\bibitem{wang2019multi}
Z.~Wang, Y.~Wu, and Q.~Niu.
\newblock Multi-sensor fusion in automated driving: A survey.
\newblock {\em Ieee Access}, 8:2847--2868, 2019.

\bibitem{36wu2017lidar}
T.-E. Wu, C.-C. Tsai, and J.-I. Guo.
\newblock Lidar/camera sensor fusion technology for pedestrian detection.
\newblock In {\em 2017 Asia-Pacific Signal and Information Processing Association Annual Summit and Conference (APSIPA ASC)}, pages 1675--1678. IEEE, 2017.

\bibitem{xu2023bridging}
R.~Xu, J.~Li, X.~Dong, H.~Yu, and J.~Ma.
\newblock Bridging the domain gap for multi-agent perception.
\newblock In {\em 2023 IEEE International Conference on Robotics and Automation (ICRA)}, pages 6035--6042. IEEE, 2023.

\bibitem{xu2022cobevt}
R.~Xu, Z.~Tu, H.~Xiang, W.~Shao, B.~Zhou, and J.~Ma.
\newblock Cobevt: Cooperative bird's eye view semantic segmentation with sparse transformers.
\newblock {\em arXiv preprint arXiv:2207.02202}, 2022.

\bibitem{v2xvitxu2022v2x}
R.~Xu, H.~Xiang, Z.~Tu, X.~Xia, M.-H. Yang, and J.~Ma.
\newblock V2x-vit: Vehicle-to-everything cooperative perception with vision transformer.
\newblock In {\em European conference on computer vision}, pages 107--124. Springer, 2022.

\bibitem{xu2022opv2v}
R.~Xu, H.~Xiang, X.~Xia, X.~Han, J.~Li, and J.~Ma.
\newblock Opv2v: An open benchmark dataset and fusion pipeline for perception with vehicle-to-vehicle communication.
\newblock In {\em 2022 International Conference on Robotics and Automation (ICRA)}, pages 2583--2589. IEEE, 2022.

\bibitem{yu2022dair}
H.~Yu, Y.~Luo, M.~Shu, Y.~Huo, Z.~Yang, Y.~Shi, Z.~Guo, H.~Li, X.~Hu, J.~Yuan, et~al.
\newblock Dair-v2x: A large-scale dataset for vehicle-infrastructure cooperative 3d object detection.
\newblock In {\em Proceedings of the IEEE/CVF Conference on Computer Vision and Pattern Recognition}, pages 21361--21370, 2022.

\bibitem{fpvrcnn2022keypoints}
Y.~Yuan, H.~Cheng, and M.~Sester.
\newblock Keypoints-based deep feature fusion for cooperative vehicle detection of autonomous driving.
\newblock {\em IEEE Robotics and Automation Letters}, 7(2):3054--3061, 2022.

\bibitem{zhao2017cooperative}
X.~Zhao, K.~Mu, F.~Hui, and C.~Prehofer.
\newblock A cooperative vehicle-infrastructure based urban driving environment perception method using a ds theory-based credibility map.
\newblock {\em Optik}, 138:407--415, 2017.

\bibitem{zhu2020deformable}
X.~Zhu, W.~Su, L.~Lu, B.~Li, X.~Wang, and J.~Dai.
\newblock Deformable detr: Deformable transformers for end-to-end object detection.
\newblock {\em arXiv preprint arXiv:2010.04159}, 2020.

\end{thebibliography}
}

\end{document}